\documentclass[10pt,twocolumn,letterpaper]{article}

\usepackage[pagenumbers]{cvpr} 

\definecolor{cvprblue}{rgb}{0.21,0.49,0.74}
\usepackage[pagebackref,breaklinks,colorlinks,allcolors=cvprblue]{hyperref}
\usepackage{graphicx}
\usepackage{amsmath}
\usepackage{amssymb}
\usepackage{booktabs}
\usepackage{multirow}
\usepackage{todonotes}

\usepackage{colortbl} 
\usepackage{arydshln}
\usepackage[dvipsnames]{xcolor}
\usepackage{pifont} 
\usepackage{tikz} 
\definecolor{tabfirst}{rgb}{1, 0.75, 0.7}
\definecolor{tabsecond}{rgb}{1, 0.85, 0.65}
\definecolor{tabthird}{rgb}{1, 0.96, 0.7}

\def\paperID{2324} 
\def\confName{CVPR}
\def\confYear{2026}
\usepackage{colortbl}
\usepackage{arydshln}
\usepackage[dvipsnames]{xcolor}
\title{Reloc-VGGT: Visual Re-localization with Geometry Grounded Transformer
}

\author{Tianchen Deng\textsuperscript{1}*, Wenhua Wu\textsuperscript{1}*, Kunzhen Wu\textsuperscript{1}*, Guangming Wang\textsuperscript{3}, Siting Zhu\textsuperscript{1},\\ Shenghai Yuan\textsuperscript{2}, Xun Chen\textsuperscript{2}, Guole Shen\textsuperscript{1}, Zhe Liu\textsuperscript{1},
 Hesheng Wang\textsuperscript{1}
 \\
{\textsuperscript{\rm 1} Shanghai Jiao Tong University}
{\textsuperscript{\rm 2} Nanyang Technological University}
{\textsuperscript{\rm 3} Cambridge University}
}

\begin{document}

\maketitle

\renewcommand{\thefootnote}{} 
\footnotetext{ The first three authors contribute equally to this paper. }

\begin{abstract}
Visual localization has traditionally been formulated as a pair-wise pose regression problem. Existing approaches mainly estimate relative poses between two images and employ a late-fusion strategy to obtain absolute pose estimates. However, late motion average is often insufficient for effectively integrating spatial information, and its accuracy degrades in complex environments. In this paper, we present the first visual localization framework that performs multi-view spatial integration through an early-fusion mechanism, enabling robust operation in both structured and unstructured environments. Our framework is built upon the VGGT backbone, which encodes multi-view 3D geometry, and we introduce a pose tokenizer and projection module to more effectively exploit spatial relationships from multiple database views.
Furthermore, we propose a novel sparse mask attention strategy that reduces computational cost by avoiding the quadratic complexity of global attention, thereby enabling real-time performance at scale. Trained on approximately eight million posed image pairs, Reloc-VGGT demonstrates strong accuracy and remarkable generalization ability. Extensive experiments across diverse public datasets consistently validate the effectiveness and efficiency of our approach, delivering high-quality camera pose estimates in real time while maintaining robustness to unseen environments. Our code and models will be publicly released upon acceptance.\href{https://github.com/dtc111111/Reloc-VGGT }{https://github.com/dtc111111/Reloc-VGGT }.

\end{abstract}

\section{Introduction}
Visual re-localization has been a fundamental challenge in robotics and computer vision, which aims to estimate query 6-degree-of-freedom (6-Dof) poses from a database posed images in a world coordinate system. This technology has broad applications in domains including autonomous driving, robotics navigation, VR, and AR, with specific importance in SLAM~\cite{plgslam,mneslam}.

\begin{figure}[!t]
    \centering
    \includegraphics[width=\linewidth]{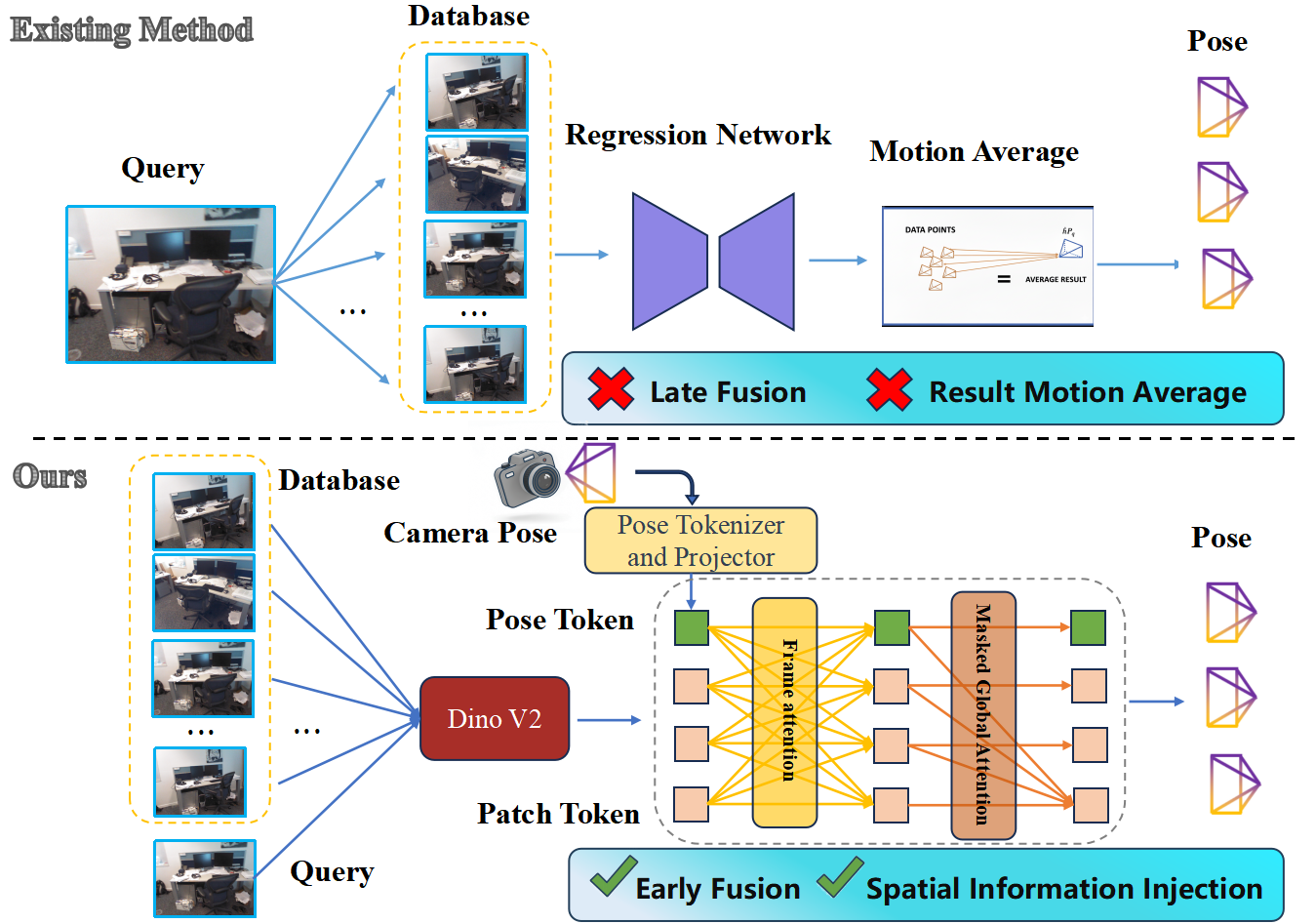}

    \caption{\textbf{Illustration of an existing visual relocalization baseline (top) and our proposed method (bottom).}  The upper pipeline illustrates conventional relocalization methods, which rely on pair-wise pose regression and employ late fusion to estimate camera poses. In contrast, our approach introduces a multi-frame relocalization framework that performs early fusion through spatial information injection, enabling more effective spatial feature integration. Our method achieves state-of-the-art performance across standard visual relocalization and pose regression benchmarks.
}

    \label{fig:teaser}
\end{figure}

Traditional visual relocalization methods rely on handcrafted feature points such as ORB~\cite{orbslam} or SIFT~\cite{vins}, or on learned features such as SuperPoint~\cite{superpoint} and SuperGlue~\cite{superglue}, to reconstruct explicit 3D maps. Camera poses are then solved via geometric optimization based on matched 2D–3D correspondences. Alternatively, some recent methods~\cite{brachmann2018learning,brachmann2021visual,brachmann2023accelerated,li2020hierarchical} avoid explicit map reconstruction and instead recover relative poses directly from neural networks, either by regressing the pose from images or by employing implicit scene representations such as NeRF~\cite{NeRF,nerf--,inerf} or 3D Gaussian Splatting~\cite{3dgs,deng2025vpgs,compactslam}. However, these approaches often suffer from limited generalization and typically require ground-truth keypoint correspondences as supervision or per-scene optimization.

Absolute Pose Regression (APR) methods~\cite{brahmbhatt2018geometry,kendall2017geometric,shavit2021learning,shavit2023coarse,walch2017image,unilgl} directly estimate the camera pose from images, providing fast inference and reasonable accuracy, but they remain highly scene-specific and require dense viewpoint coverage during training, which restricts their practical deployment. Relative Pose Regression (RPR)~\cite{arnold2022map,balntas2018relocnet,zhou2020learn} offers a more flexible alternative by predicting the relative pose between a query and a set of database images. These methods avoid per-scene retraining and enable fast execution, but their accuracy remains limited. While several RPR-based methods~\cite{ding2019camnet,arnold2022map} generalize across datasets, this generalization often comes at the cost of further degraded pose accuracy.

Reloc3R~\cite{reloc3r} is the first relocalization framework built upon a 3D foundation model. Based on the DUST3R architecture~\cite{dust3r}, it introduces a motion-averaging strategy to recover absolute poses and achieves strong performance across novel scenes, in both accuracy and test-time efficiency. However, it overlooks multi-frame spatial information by fusing views only through simple motion averaging, which fails to capture geometric relationships under challenging motion or limited viewpoint overlap. 

Motivated by this gap, we propose a novel multi-view relocalization framework built on VGGT~\cite{vggt}. First, we introduce a source frame pose tokenizer and projection module to inject spatial relative pose information from multiple reference frames into the network. The injection of pose tokens enables better interaction with the patch and register tokens within the alternating attention block in early fusion, allowing more effective utilization of spatial information, thereby improving pose estimation accuracy.
 Second, to accelerate test-time inference, we replace the full global attention in VGGT with a sparse mask attention mechanism that retains only the attention between the query and selected anchor frames. This reduces the quadratic attention complexity to linear form ($\mathcal{O}(N^2) \rightarrow \mathcal{O}(5N-5)$), greatly boosting inference speed and enabling scalable relocalization over long sequences and large environments.

Overall, our contributions are shown as follows:
\begin{itemize}
    \item We propose a novel multi-frame visual localization method with 3D foundation model, enabling universal generalization across diverse environments, and high camera pose accuracy.
    \item A novel pose tokenizer and projection module are designed to align the 3D pose token with 2D patch token and better leverage the relative spatial information from multi-view during the attention block.  
\item We propose a novel sparse mask attention method to further enhance test-time inference speed and reduce memory requirements by reducing the computational complexity from quadratic to linear. Comprehensive experiments across different datasets consistently demonstrate the effectiveness of our proposed framework.

\end{itemize}

\begin{figure*}[h]
    \centering
    \includegraphics[width=\linewidth]{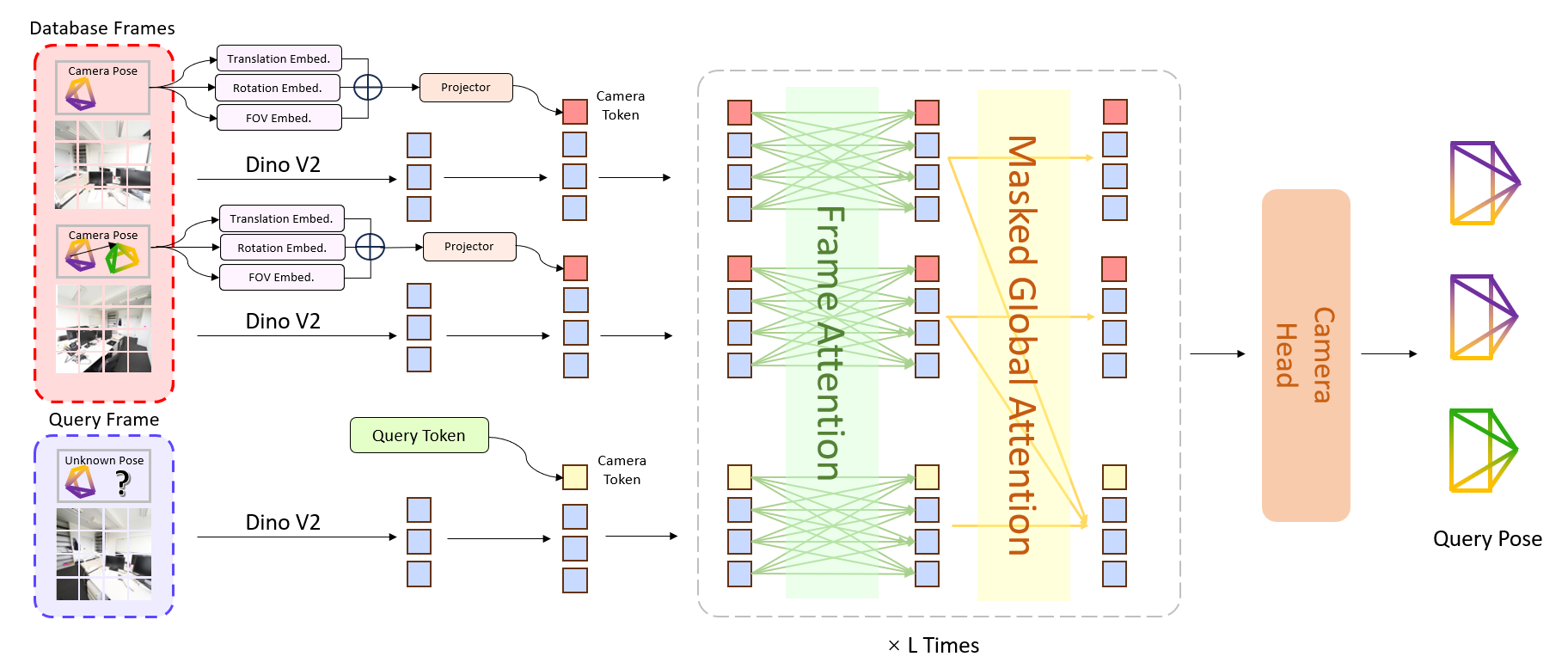}
    \vspace{-0.2cm}
    \caption{\textbf{System Overview.} We propose a novel visual re-localization framework with 3D foundation model. We introduce a novel pose tokenizer and projection module to better leverage the spatial
     information for early fusion and design a sparse mask attention strategy to enhance test-time inference speed and reduce computational complexity from quadratic to linear.}

    \label{fig:system}
    \vspace{-0.2cm}
\end{figure*}

\section{Related work}
Visual relocalization has advanced rapidly and found applications in SLAM~\cite{deng2025mcnslam,grsslam,neslam,mgslam}, navigation~\cite{lenav}, autonomous driving~\cite{prosgnerf,GaussianDWM}, VR~\cite{incremental}. Here, we review related work on 3D-map based localization, pose regression, and foundation model.  

\noindent \textbf{3D-map based Localization}
Traditional 3D-map based methods rely on explicit geometric reconstruction to establish 2D–3D correspondences for pose estimation. Early approaches leverage handcrafted features (e.g., ORB~\cite{orbslam}, SIFT~\cite{vins}) or learned local descriptors (e.g., SuperPoint~\cite{superpoint}, SuperGlue~\cite{superglue}) to match query images with pre-built 3D maps, followed by PnP or bundle adjustment. However, these methods suffer from poor generalization to unseen environments and high computational costs for map construction. Recent implicit scene representations have emerged to avoid explicit map storage. 
iNeRF~\cite{inerf} pioneered the use of Neural Radiance Fields (NeRF)~\cite{NeRF} for camera relocalization, achieving pose optimization by minimizing the photometric error between query images and corresponding renders from a pre-trained NeRF.
Lens~\cite{LENS} further enhanced localization accuracy by leveraging NeRF to synthesize additional training data.
Building on this line of work, NeRF-SCR~\cite{chen2024leveraging} introduced uncertainty analysis to improve data efficiency and prevent contamination from low-quality generated samples.
More recently, SplatLoc~\cite{zhai2025splatloc} adopted 3D Gaussian Splatting (3DGS)~\cite{3dgs} as its scene representation and designed a specialized descriptor decoder for Gaussian primitives, achieving remarkable localization performance.
However, these methods require per-scene optimization, which limits their applicability in real-time scenarios.

\noindent \textbf{Pose Regression}
Pose regression methods directly estimate camera poses from images without explicit 3D maps, divided into Absolute Pose Regression (APR) and Relative Pose Regression (RPR).
APR methods directly regress 6-DoF camera poses from query images. PoseNet~\cite{kendall2015posenet} pioneers the use of convolutional neural networks to predict camera poses directly from individual images.Subsequently, VidLoc~\cite{clark2017vidloc} extends to sequential images, leveraging temporal continuity to achieve enhanced pose estimation accuracy and smoothness. Further advancing this direction, LSG~\cite{xue2019local} incorporates relative pose constraints from visual odometry to refine camera tracking. 
However, these methods exhibit inherent limitation in scene generalization, typically requiring retraining for new scenes.
RPR methods predict relative poses between query and reference images, offering enhanced flexibility.
RelocNet~\cite{balntas2018relocnet} introduces feature descriptors learned through nearest-neighbor matching and continuous metric learning to achieve relative pose regression between query images and their nearest neighbors.
CamNet~\cite{ding2019camnet} proposes a coarse-to-fine learning framework that progressively refines pose estimation accuracy.
Relpose-GNN~\cite{turkoglu2021visual} constructs graph structures connecting query and reference images, enabling efficient feature propagation to obtain consistent camera poses.
Despite demonstrating better generalization capabilities than APR methods, these RPR approaches still require dataset-specific training and thus remain limited in scalability.

\noindent \textbf{Foundation Model} The Transformer architecture~\cite{vaswani2017attention} has been widely adopted in large-scale vision models~\cite{bai2024sequential, kirillov2023segment,peebles2023scalable} and language models~\cite{bai2023qwen, chowdhery2023palm, minaee2024large}. With the advancement of large-scale pretrained models, 3D vision foundation models~\cite{deng2025best3dscenerepresentation} have attracted significant research attention. DUSt3R~\cite{dust3r} introduces a novel paradigm for 3D reconstruction from image collections without requiring camera poses or calibration parameters, unifying various 3D vision tasks under a single framework. MUSt3R~\cite{cabon2025must3r} extends the pairwise formulation to multi-view scenarios and incorporates a multi-layer memory mechanism, substantially improving reconstruction efficiency. VGGT~\cite{vggt} presents an elegantly designed feed-forward network that directly infers all essential 3D attributes from image collections with remarkable efficiency. Notably, Reloc3R~\cite{reloc3r} pioneers the application of such models to visual relocalization tasks. However, its multi-view fusion relies solely on simple averaging strategies, failing to fully exploit inter-frame geometric relationships. In contrast, our method leverages relative pose token injection and sparse attention mechanisms to comprehensively capture geometric correlations between frames and maintain real-time performance.

\section{Method}
We propose Reloc-VGGT, a universal visual re-localization framework. The input to our method consists of query and source images $\{\mathbf{I_i}\}$ from database and the relative 6DOF camera poses$\{R_i,T_i\}$ of source images. The output of our framework is the estimated poses to the world coordinate system defined by the database images. Fig.~\ref{fig:system} illustrates an overview of the proposed framework.  It is composed of two main modules:(i) relative pose regression (ii) pose tokenizer and projector(Sec.~\ref{Sec:tokenizer});
(iii) sparse mask attention (Sec.~\ref{Sec:mask attention});
We elaborate on the entire pipeline of our system in the following subsections.

\subsection{Relative Pose Regression}
For each query image $I_q$, we retrieve the top-$K$ nearest database images using an off-the-shelf image retrieval method~\cite{netvlad}, forming a set of source–query pairs $\mathbf{Q} = {(I_{d_k}, I_q) \mid k=1, \ldots, K}$. The pose of the query image is then estimated based on the known poses of these source images. Existing methods, such as Reloc3R~\cite{reloc3r}, typically perform pairwise matching between the query and each source image and then fuse the results by averaging the predicted motions. In contrast, our approach performs multi-frame matching, enabling more effective aggregation of spatial information through alternating frame attention and global attention.

We adopt VGGT~\cite{vggt} as the 3D backbone of our relative pose regression network. Given the query image, the $K$ retrieved source images, and their corresponding known poses, the network directly predicts the pose of the query image. Formally, the process can be expressed as:

\begin{equation}
\{R,T\}_q = f_{\theta}(I_q, \{(I_{d_k}, \{R,T\}_{d_k})\}_{k=1}^{K}),
\end{equation}
where $\{R,T\}_{d_k}$ denotes the known pose of the $k$-th source image, and $\{R,T\}_q$ is the predicted pose of the query image.
For the camera parameters $\{R, T\}$, we follow the parametrization from~\cite{vggt}, where the rotation matrix is represented using a quaternion, and an additional field-of-view (FoV) term is included. These components are then concatenated into a single vector, consisting of the rotation quaternion $\mathbf{q} \in \mathbb{R}^4$, the translation vector $\mathbf{t} \in \mathbb{R}^3$, and the field-of-view parameter $\mathbf{f} \in \mathbb{R}^2$.
We assume that the camera's principal point is at the image center, which is common in SfM frameworks~\cite{schonberger2016structure}.

For the query frame within the input sequence, we explore two different placement strategies. The first is to treat the query frame as the anchor frame, leveraging VGGT’s design where the first frame is interpreted as the world coordinate system. In this case, the final absolute pose can be obtained directly by motion averaging. However, this approach makes it impossible to inject relative pose tokens, since the pose of the query frame is unknown at inference time. Alternatively, since VGGT treats all non-anchor frames symmetrically, we position the query frame at the last index of the sequence and select the most similar database frame—identified using visual place recognition methods~\cite{netvlad,unipr-3d}-as the anchor frame. This setup allows us to encode and inject relative pose tokens for all other source frames, including those associated with the query.
\subsection{Pose Tokenizer and Projector}
\label{Sec:tokenizer}
Unlike previous pose regression methods that rely on late fusion to combine source image poses, we adopt an intermediate fusion strategy. Specifically, we encode the relative pose between the query and source images and align it with the register and patch tokens before feeding them into the alternating attention blocks. This design enables more effective spatial reasoning and feature interaction across multiple frames.

Building upon this, we apply learnable Fourier embeddings~\cite{NeRF} to encode the relative translation vector, the rotation quaternion, and the field-of-view (FoV) parameters:

 \begin{equation}
\begin{aligned}
\gamma(\mathbf{t}) &= 
\big[
\sin(2^{l}\pi \mathbf{t}),\;
\cos(2^{l}\pi \mathbf{t})
\big]_{l=0}^{L_t-1}, \\
\gamma(\mathbf{q}) &= 
\big[
\sin(2^{l}\pi \mathbf{q}),\;
\cos(2^{l}\pi \mathbf{q})
\big]_{l=0}^{L_d-1}, \\
\gamma(\mathbf{f}) &= 
\big[
\sin(2^{l}\pi \mathbf{f}),\;
\cos(2^{l}\pi \mathbf{f})
\big]_{l=0}^{L_f-1}.
\end{aligned}
 \end{equation}
We set $L_t=10,L_d=4, L_f=4$.
Finally, we introduce a camera token  projector, implemented as an MLP, to align the register tokens, 2D patch tokens with the 3D camera tokens, thereby enhancing the spatial consistency of sequence-level representations.
\begin{equation}
\mathbf{f}_{cam} = \mathrm{MLP}_{\text{cam}}\!\left(\gamma(\mathbf{x}),\,\gamma(\mathbf{q}), \gamma(\mathbf{f})\right)
\label{eq:pose-projector}
\end{equation}

\begin{figure*}[h]
    \centering
    \includegraphics[width=\linewidth]{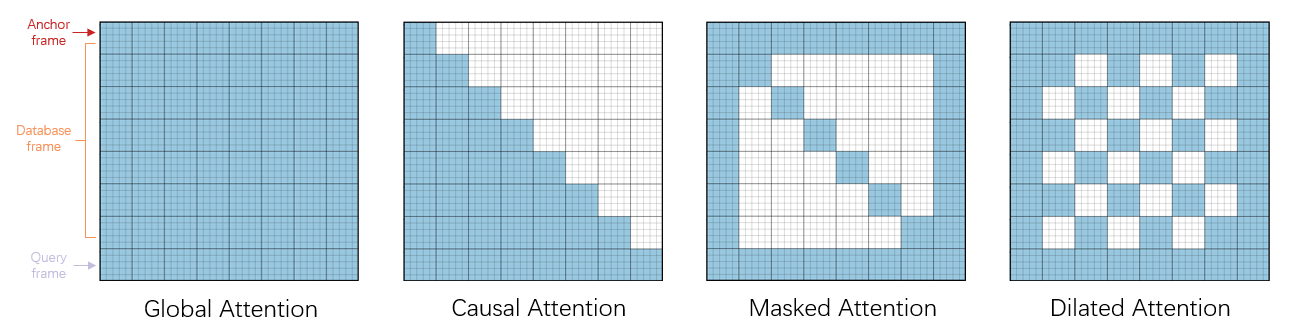}
    \vspace{-0.4cm}
    \caption{We visualize the global attention maps under different masking strategies. From left to right, the figures correspond to the original global attention in VGGT, the causal attention in StreamVGGT~\cite{streamvggt}, and our proposed sparse mask attention and dilated mask attention.
}
    \label{fig:maskattention}
    \vspace{-0.4cm}
\end{figure*}

\subsection{Sparse Mask Attention}
\label{Sec:mask attention}
In the original VGGT architecture, the alternating attention framework consists of two types of blocks: frame attention and global attention. While this design achieves highly accurate pose estimation, the global attention module introduces significant computational overhead with a quadratic complexity of $\mathcal{O}(n^2)$—particularly problematic when the sequence length increases. Recent works such as FastVGGT~\cite{fastvggt} have analyzed attention maps and revealed substantial redundancy in global attention, uncovering a key insight: attention patterns across tokens exhibit a high degree of similarity, resulting in large portions of unnecessary computation, a phenomenon also observed in models like DINOV2~\cite{dinov2}.

Motivated by these observations, we propose a task-specific sparse mask attention mechanis tailored for visual re-localization. As illustrated in Fig.~\ref{fig:maskattention}, we compare the original global attention in VGGT, the masked attention in StreamVGGT~\cite{streamvggt}, and our proposed sparse mask attention. Our method significantly reduces redundant attention computation while maintaining accuracy, making it feasible to perform multi-frame relocalization efficiently at large scales.

\noindent \textbf{Design Rationale.}
In the VGGT architecture, the first frame is defined as the world coordinate system, with all tokens registered relative to this reference frame. This setup, along with the subsequent training pipeline, makes the first frame a crucial anchor for maintaining spatial consistency throughout the sequence. Therefore, we preserve all attention connections associated with the anchor frame.
Second, as shown in the second illustration of Fig.~\ref{fig:maskattention}, StreamVGGT adopts a causal mask because its sequential streaming input allows each frame to attend only to previous frames. However, our task only requires estimating the pose of the query frame, so we retain all attention interactions involving the query frame as well. This corresponds to the structure shown in Fig.~\ref{fig:maskattention}.

Beyond these interactions, we further find a dilated attention mask for the remaining source frames to sparsify computation while still preserving the necessary spatial context. Nevertheless, this approach still incurs substantial computation and relatively slow inference, and its performance is inferior to the proposed sparse mask attention mechanism.

To this end, we propose the proposed sparse mask attention strategy and introduce it into the global attention block. It reduces the quadratic attention complexity to linear form ($\mathcal{O}(N^2) \rightarrow \mathcal{O}(5N-5)$), transforming global attention into a scalable component suitable for long-sequence relocalization.

\subsection{Training Loss}
The output of our network consists of three components:  
1) the rotation quaternion, which captures the relative change in camera orientation,  
2) the translation vector, which represents the change in camera position, and  
3) the field-of-view (FoV), which encodes the camera's intrinsic parameters.  To supervise these outputs, we adopt an $\mathcal{L}_1$ loss function. Since translation is sensitive to scene scale, we apply a scale-aware transformation to the translation vector in the loss:

\begin{equation}
\mathcal{L}_{\text {pose }}=\sum_{i=0}^{K}\left(\left\|\hat{\mathbf{q}_i}-\mathbf{q}_i\right\|+\left\|\hat{\mathbf{t}_i}-\mathbf{t}_i\right\|+\left\|\hat{\mathbf{f}_i}-\mathbf{f}_i\right\|\right)
\end{equation}
Jointly supervising all three components stabilizes training and allows the model to converge more rapidly.
We also use the rotation loss from~\cite{reloc3r} to enhance the learning of rotation components $\mathcal{L}_R= \arccos \left(\frac{\operatorname{tr}\left(\hat{R}^{-1} R\right)-1}{2}\right)$.
Our training process follows a two-stage strategy. In the first stage, we freeze the ViT encoder and decoder, and only train the pose tokenizer and projector. This ensures that the injected pose tokens are properly aligned with the original register and patch tokens. In the second stage, we unfreeze the decoder and attention blocks and proceed with end-to-end training.

\section{Experiments}
In this section, we first describe the implementation details, training datasets, and baseline methods used for comparison. We then present results across various test datasets, covering both two-frame pairwise relative pose estimation and full visual relocalization experiments. Finally, we conduct ablation studies and additional analyses to evaluate the effectiveness of each component in our framework.

\subsection{Implementation Details}
Our model architecture follows VGGT~\cite{vggt} with $L = 24$ alternating frame and global attention layers. To accelerate inference, we incorporate FlashAttention-2~\cite{flashattention}. The model is initialized using pre-trained VGGT weights. We train using the AdamW optimizer with a hybrid learning rate schedule: a linear warm-up over the first 0.5 epochs, followed by cosine decay, with a peak learning rate of $1\times 10^{-6}$. All experiments are conducted on 4 NVIDIA A100 GPUs.

For the visual localization task, following Reloc3R~\cite{reloc3r}, we adopt NetVLAD~\cite{netvlad} for image retrieval and select the top 10 most similar image pairs without applying distance-based clustering.

\subsection{Dataset and Baseline}
The choice of training datasets follows Reloc3R~\cite{reloc3r}. Each input image is center-cropped based on its principal point and resized to a fixed width of 518 pixels. For the pair-wise relative camera pose estimation experiments, we evaluate on the ScanNet1500~\cite{scannet} datasets,  Following previous research~\cite{reloc3r}, we employ three metrics in pair-wise experiments: AUC@5/10/20. These
metrics calculate the area under the curve of pose accuracy
using thresholds of $\tau= 5/10/20$ degrees for the minimum
of rotation and translation angular errors. For multi view relative pose experiments, we conduct the evaluation on the CO3Dv2~\cite{co3dv2} dataset. We use the metrics in~\cite{posediffusion,dust3r} RRA@15  the relative rotation
accuracy within 15 degrees, and RTA@15 for the relative translation accuracy within 15 degrees.  the mean
average accuracy (mAA@30, also called AUC@30).
To evaluate the visual re-localization experiments, we conduct evaluations on two public benchmarks that cover both indoor and outdoor environments: the 7-Scenes dataset~\cite{7scenes} and the Cambridge Landmarks dataset~\cite{cambridge}. or each scene,
we report the median translation and rotation errors (in meters and degrees, respectively). 

\begin{table}[htbp]
\centering
\scalebox{0.78}{
\setlength{\tabcolsep}{0.5mm}{
\begin{tabular}{c|l|ccc|c}
\toprule
& \textbf{Methods} & \multicolumn{3}{c|}{\textbf{ScanNet1500}} & \textbf{Inference time} \\
& & AUC@5 & AUC@10 & AUC@20 & \\
\midrule
\multirow{5}{*}{\textbf{\rotatebox{90}{Non-PR}}} 
& Efficient LoFTR ~\cite{LoFTR} & 19.20 & 37.00 & 53.60 & 40 ms \\
& ROMA ~\cite{roma} & 28.90 & 50.40 & 68.30 & 300 ms \\
& DUSt3R ~\cite{dust3r} & 23.81 & 45.91 & 65.57 & 441 ms \\
& MASt3R ~\cite{cabon2025must3r} & 28.01 & 50.24 & 68.83 & 294 ms \\
& NoPoSplat ~\cite{noposplat} & 31.80 & 53.80 & 71.70 & $>$2000 ms \\
\midrule
\multirow{7}{*}{\rotatebox{90}{\textbf{PR}}} 
& Map-free (Regress-SN) ~\cite{arnold2022map} & 1.84 & 8.75 & 25.33 & 10 ms \\
& Map-free (Regress-MF) ~\cite{arnold2022map} & 0.50 & 3.48 & 13.15 & 10 ms \\
& ExReNet (SN) ~\cite{ExReNet} & 2.30 & 10.71 & 26.13 & 17 ms \\
& ExReNet (SUNCG) ~\cite{ExReNet} & 1.61 & 7.00 & 18.03 & 17 ms \\
& Reloc3r-512 ~\cite{reloc3r} & 34.79 & 58.37 & 75.56 & 25 ms \\
& \textbf{Reloc-VGGT} & \textbf{36.35} & \textbf{58.62} & \textbf{75.90} & 45 ms \\
\bottomrule
\end{tabular}}}
\caption{Pair-wise relative pose estimation experiments on the ScanNet1500 dataset~\cite{scannet}.}
\label{tab:pairwise}
\end{table}

\begin{table}[htbp]
\centering
\scalebox{0.90}{
\setlength{\tabcolsep}{1mm}{
\begin{tabular}{l|l|ccc}
\toprule
 & \textbf{Methods} & \textbf{RRA@15} & \textbf{RTA@15} & \textbf{mAA@30} \\
\midrule
\multirow{8}{*}{\rotatebox{90}{Non-PR}} 
& PixSfM~\cite{pixsfm} & 33.7 & 32.9 & 30.1 \\
& RelPose~\cite{relpose} & 57.1 & - & - \\
& PoseDiffusion~\cite{posediffusion} & 80.5 & 79.8 & 66.5 \\
& RelPose++~\cite{relposepp} & 82.3 & 77.2 & 65.1 \\
& RayDiffusion*~\cite{raydiffusion} & 93.3 & - & - \\
& VGGSFm~\cite{vggsfm} & 92.1 & 88.3 & 74.0 \\
& DUS3R (w/ PnP)~\cite{dust3r} & 94.3 & 88.4 & 77.2 \\
& MASt3R~\cite{mast3r} & 94.6 & 91.9 & 81.1 \\
\midrule
\multirow{4}{*}{\rotatebox{90}{PR}} 
& PoseReg~\cite{posediffusion} & 53.2 & 49.1 & 45.0 \\
& RayReg*~\cite{raydiffusion} & 89.2 & - & - \\
& Reloc3r-512 ~\cite{reloc3r} & 95.8 & 93.7 & 82.9 \\
& \textbf{Reloc-VGGT} & \textbf{96.1} & \textbf{94.5} & \textbf{83.4} \\
\bottomrule
\end{tabular}}}
\caption{Multi-frame relative pose evaluation the CO3Dv2 dataset~\cite{co3dv2}.}
\label{tab:multiview}
\end{table}

\subsection{Relative Camera Pose Estimation}
\noindent\textbf{Pair-wise relative pose}
In this section, we evaluate the accuracy of pair-wise relative pose estimation. The results are shown in Tab.~\ref{tab:pairwise}. We compare our method with several state-of-the-art approaches, including Dust3R~\cite{dust3r}, NopoSplat~\cite{noposplat}, Map-Free~\cite{arnold2022map}, and ExReNet~\cite{ExReNet}. Experimental results demonstrate that our method achieves competitive performance in two-view matching and pose estimation tasks.

\noindent\textbf{Multi-view relative pose}
The multi-view pose estimation results are shown in Tab.~\ref{tab:multiview}. We compare Reloc3r with recent SfM-based approaches, PixSfM~\cite{pixsfm} and VGGSfM~\cite{vggsfm}, as well as data-driven methods including RelPose~\cite{relpose}, RelPose++~\cite{relposepp}, PoseDiffusion~\cite{posediffusion}, RayDiffusion~\cite{raydiffusion}, DUSt3R (with PnP)~\cite{dust3r}, and MASt3R~\cite{mast3r}.
Unlike Reloc3r, which aggregates pairwise predictions, our method more effectively integrates information across multiple frames, enabling consistent multi-view matching and robust joint pose estimation. This richer temporal and cross-view fusion leads to improved geometric consistency and strengthens pose recovery in challenging multi-view scenarios.
\subsection{Visual Re-localization}
In this section, we evaluate the absolute pose estimation capability of our approach.
Tab.~\ref{tab:7scenes} reports results on the indoor 7-Scenes dataset~\cite{7scenes}. We compare our method against state-of-the-art absolute pose regression (APR) and relative pose regression (RPR) approaches. Following the evaluation protocol of Reloc3R, we additionally distinguish between seen and unseen scenes, where the latter contains entirely novel environments during testing.
Across both settings, our method consistently achieves state-of-the-art performance in terms of translation and rotation errors. In unseen scenes, the performance gap becomes even more pronounced, demonstrating the strong generalization ability of our design. These results clearly validate the effectiveness of our early fusion module, which integrates relative pose cues from the input sequence to produce more accurate and stable absolute pose estimates. Moreover, compared with APR methods, our approach attains superior accuracy without any scene-specific training, highlighting its robustness and adaptability across diverse environments.

\begin{figure*}[h]
    \centering
    \includegraphics[width=\linewidth]{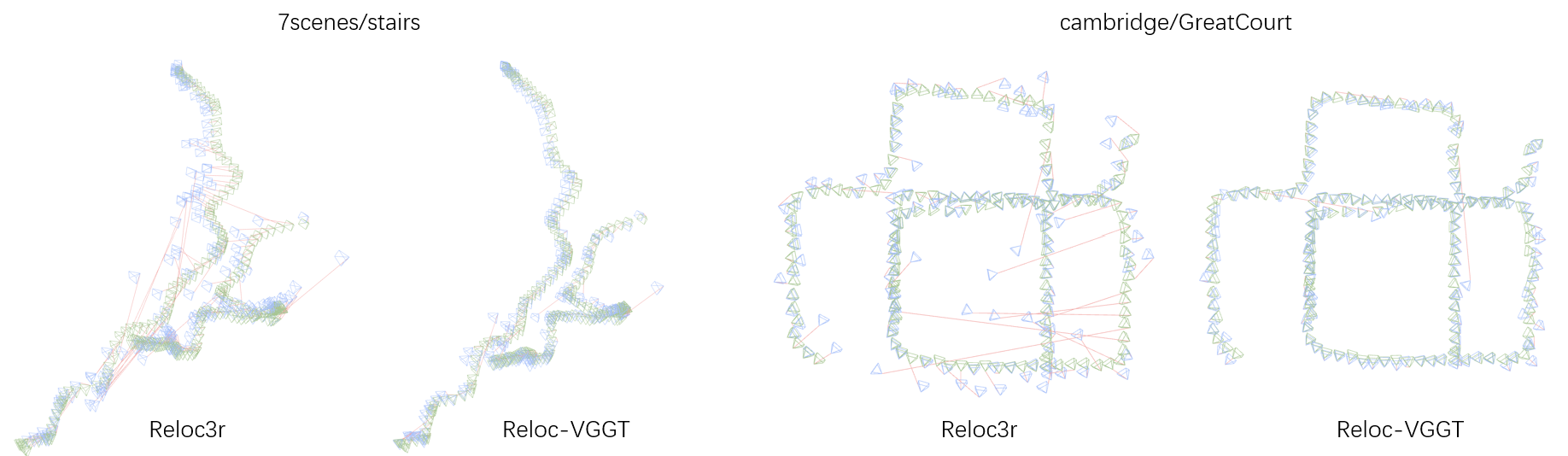}
    \vspace{-0.3cm}
    \caption{We visualize pose estimates for two scenes: stairs from the 7 Scenes dataset~\cite{7scenes} and GreatCourt from Cambridge Landmarks
~\cite{cambridge}. We compare our results with the exisitng sota method Reloc3R~\cite{reloc3r}.}
    \label{fig:visualization}
    \vspace{-0.3cm}
\end{figure*}

Tab.~\ref{tab:cambridge} further presents results on the outdoor Cambridge Landmarks dataset~\cite{oxfordrobo}, a widely used benchmark for regression-based localization. In this more challenging outdoor setting, all RPR methods—under both seen and unseen configurations—experience notable performance degradation. Nevertheless, our method still achieves a significant improvement over existing approaches, including the foundation-model-based Reloc3R, yielding substantially lower translation and rotation errors. This confirms the strength of our overall architecture, where the pose tokenizer, pose projector, and the associated early fusion and spatial information injection collectively contribute to robust global pose estimation.
Although the introduction of sparse mask attention brings a slight reduction in accuracy, it achieves a favorable balance between accuracy and computational efficiency, making the model more suitable for real-world deployment.
Visual comparisons are provided in Fig.~\ref{fig:visualization}.

\begin{table*}[t]
\centering
\scalebox{0.78}{
\small
\setlength{\tabcolsep}{0.6mm}{
\begin{tabular}{lcccccccccc}
\toprule
\textbf{Methods} & \textbf{Chess} & \textbf{Fire} & \textbf{Heads} & \textbf{Office} & \textbf{Pumpkin} & \textbf{RedKitchen} & \textbf{Stairs} & \textbf{Average} & \textbf{Dataset-specific training time} \\
\midrule
\multicolumn{10}{l}{\textbf{APR}} \\
LENS~\cite{LENS} & 0.03 / 1.30 & 0.10 / 3.70 & 0.07 / 5.80 & 0.07 / 1.90 & 0.08 / 2.20 & 0.09 / 2.20 & 0.14 / 3.60 & 0.08 / 3.00 & Days / scene \\
PMNet~\cite{PMNet} & 0.03 / 1.26 & 0.04 / 1.76 & 0.06 / 1.68 & 0.06 / 1.90 & 0.07 / 1.91 & 0.08 / 2.23 & 0.11 / 2.97 & 0.06 / 1.96 & Days / scene \\
DFNet~\cite{DFNet}+NeFeS~\cite{NeFeS} & \textbf{0.02} / \textbf{0.57} & \textbf{0.02} / \textbf{0.74} & \textbf{0.02} / \textbf{1.28} & \textbf{0.02} / \textbf{0.56} & \textbf{0.02} / \textbf{0.55} & \textbf{0.02} / \textbf{0.57} & \textbf{0.05} / \textbf{1.28} & \textbf{0.02} / \textbf{0.79} & Days / scene \\
Marepo~\cite{Marepo} & \textbf{0.02} / 1.24 & \textbf{0.02} / 1.39 & \textbf{0.02} / 2.03 & 0.03 / 1.26 & 0.04 / 1.48 & 0.04 / 1.71 & 0.06 / 1.67 & 0.03 / 1.54 & 15min / scene \\
\midrule
\multicolumn{10}{l}{\textbf{RPR (Seen)}} \\
EssNet (7S)~\cite{zhou2020learn} & -- & -- & -- & -- & -- & -- & -- & 0.22 / 8.03 & Hours \\
Relative PN (7S)~\cite{RelativePN} & 0.13 / 6.46 & 0.26 / 12.72 & 0.14 / 12.34 & 0.21 / 7.35 & 0.24 / 6.35 & 0.24 / 8.03 & 0.27 / 11.82 & 0.21 / 9.30 & Hours \\
NC-EssNet (7S)~\cite{zhou2020learn} & 0.26 / 10.44 & -- & 0.26 / 10.4 & 0.18 / 5.32 & -- & 0.23 / 7.45 & -- & 0.21 / 7.50 & Hours \\
RelocNet (7S)~\cite{balntas2018relocnet} & 0.12 / 4.14 & 0.21 / 7.50 & 0.13 / 8.70 & 0.15 / 4.50 & 0.13 / 5.30 & 0.21 / 5.08 & 0.28 / 7.53 & 0.16 / 5.73 & Hours \\
Relpose-GNN~\cite{brachmann2021visual} & 0.08 / 2.70 & 0.21 / 5.30 & 0.13 / 8.70 & 0.15 / 4.30 & 0.13 / 5.09 & 0.19 / 5.30 & 0.22 / 6.50 & 0.16 / 5.20 & Hours \\
AnchorNet~\cite{anchornet} & 0.06 / 3.89 & 0.15 / 10.3 & 0.08 / 10.9 & 0.09 / 4.90 & 0.20 / 2.97 & 0.08 / 4.68 & 0.10 / 9.26 & 0.09 / 6.74 & Hours \\
CamNet~\cite{ding2019camnet} & \textbf{0.04} / \textbf{1.73} & \textbf{0.03} / \textbf{1.74} & \textbf{0.05} / \textbf{1.98} & \textbf{0.04} / \textbf{1.62} & \textbf{0.04} / \textbf{1.64} & \textbf{0.04} / \textbf{1.63} & \textbf{0.04} / \textbf{1.51} & \textbf{0.04} / \textbf{1.69} & Hours \\
\midrule
\multicolumn{10}{l}{\textbf{RPR (Unseen)}} \\
EssNet (CL)~\cite{zhou2020learn} & -- & -- & -- & -- & -- & -- & -- & 0.57 / 80.06 & None \\
NC-EssNet (CL)~\cite{zhou2020learn} & -- & -- & -- & -- & -- & -- & -- & 0.48 / 32.97 & None \\
Relative PN (U)~\cite{RelativePN} & 0.31 / 15.05 & 0.40 / 19.00 & 0.24 / 22.15 & 0.38 / 14.14 & 0.44 / 18.24 & 0.41 / 16.51 & 0.35 / 23.55 & 0.36 / 18.38 & None \\
RelocNet (SN)~\cite{balntas2018relocnet} & 0.21 / 10.9 & 0.28 / 11.4 & 0.23 / 11.8 & 0.31 / 10.3 & 0.40 / 10.9 & 0.33 / 11.4 & 0.33 / 11.4 & 0.29 / 11.3 & None \\
ImageNet+NCM ~\cite{zhou2020learn}  & -- & -- & -- & -- & -- & -- & -- & 0.19 / 4.30 & None \\
Map-free (Match)~\cite{arnold2022map} & 0.10 / 2.93 & 0.12 / 4.95 & 0.11 / 5.44 & 0.12 / 3.77 & 0.14 / 3.45 & 0.14 / 4.45 & 0.14 / 4.50 & 0.12 / 4.07 & None \\
Map-free (Regress)~\cite{arnold2022map} & 0.09 / 2.64 & 0.13 / 4.54 & 0.11 / 4.81 & 0.11 / 3.77 & 0.13 / 3.11 & 0.13 / 4.11 & 0.13 / 4.70 & 0.12 / 3.95 & None \\
ExReNet (SN)~\cite{ExReNet} & 0.06 / 3.30 & 0.09 / 5.20 & 0.09 / 3.04 & 0.07 / 2.17 & 0.11 / 2.75 & 0.09 / 3.47 & 0.09 / 7.74 & 0.09 / 3.95 & None \\
ExReNet (SUNCG)~\cite{ExReNet} & 0.05 / 2.75 & 0.07 / 2.60 & 0.08 / 3.00 & 0.07 / 2.17 & 0.09 / 2.75 & 0.07 / 3.47 & 0.07 / 7.74 & 0.07 / 3.07 & None \\
Reloc3r-512~\cite{reloc3r} & 0.027 / 0.882 & 0.028 / \textbf{0.805} & \textbf{0.013} / 0.953 & 0.041 / 0.876 & 0.062 / 1.105 & 0.043 / 1.264 & 0.070 / 1.357 & 0.041 / 1.035 & None \\
\textbf{Reloc-VGGT (mask)} & 0.027 / 0.861 & 0.028 / 0.871 & 0.014 / 0.818 & 0.039 / 0.850 & 0.053 / 1.213 & 0.042 / 1.480 & 0.074 / 1.178 & 0.039 / 1.033 & None \\
\textbf{Reloc-VGGT} & \textbf{0.025} / \textbf{0.767} & \textbf{0.024} / 0.833 & \textbf{0.013} / \textbf{0.770} & \textbf{0.036} / \textbf{0.827} & \textbf{0.048} / \textbf{1.040} & \textbf{0.039} / \textbf{1.249} & \textbf{0.038} / \textbf{0.776} & \textbf{0.031} / \textbf{0.896} & None \\
\bottomrule
\end{tabular}}}

\caption{Pose estimation results (median translation / rotation error) on the 7-Scenes dataset~\cite{7scenes}. APR: absolute pose regression; RPR: relative pose regression.}
\label{tab:7scenes}

\end{table*}

\begin{table*}[t]
\centering
\scalebox{0.78}{
\setlength{\tabcolsep}{0.6mm}{
\begin{tabular}{lcccccccccc}
\toprule
\textbf{Methods} & \textbf{GreatCourt} & \textbf{KingsCollege} & \textbf{OldHospital} & \textbf{ShopFacade} & \textbf{StMarysChurch} & \textbf{Average (4)} & \textbf{Average} & \textbf{Dataset-specific training time} \\
\midrule
\multicolumn{9}{l}{\textbf{APR}} \\
LENS~\cite{LENS} & -- & 0.33 / \textbf{0.50} & 0.44 / 0.90 & 0.27 / 1.60 & 0.53 / 1.60 & 0.39 / 1.20 & -- & Days / scene \\
PMNet~\cite{PMNet} & -- & 0.31 / 0.55 & 0.44 / 0.79 & 0.17 / 0.86 & 0.31 / 0.96 & 0.31 / 0.79 & -- & Days / scene \\
DFNet~\cite{DFNet}+NeFeS~\cite{NeFeS} & -- & 0.37 / 0.54 & 0.52 / 0.88 & \textbf{0.15} / \textbf{0.53} & 0.37 / 1.14 & \textbf{0.35} / \textbf{0.77} & -- & Days / scene \\
\midrule
\multicolumn{9}{l}{\textbf{RPR (Seen)}} \\
EssNet (CL)~\cite{zhou2020learn} & 3.20 / 2.20 & 0.48 / 1.00 & 1.14 / 2.50 & 0.48 / 2.50 & 1.52 / 3.20 & 1.08 / 3.41 & 1.37 / 2.30 & Hours \\
Relpose-GNN~\cite{brachmann2021visual} & -- & -- & -- & -- & -- & 0.91 / 2.30 & 0.85 / 2.82 & Hours \\
NC-EssNet~\cite{zhou2020learn} (CL) & -- & -- & -- & -- & -- & -- & -- & Hours \\
AnchorNet~\cite{anchornet} & -- & 0.57 / 0.88 & 1.21 / 2.55 & 0.52 / 2.27 & 1.04 / 2.69 & 0.84 / 2.10 & -- & Hours \\
\midrule
\multicolumn{9}{l}{\textbf{RPR (Unseen)}} \\
EssNet (7S)~\cite{zhou2020learn} & -- & -- & -- & -- & -- & 10.36 / 85.75 & -- & None \\
NC-EssNet (7S)~\cite{zhou2020learn} & -- & -- & -- & -- & -- & 9.78 / 24.35 & -- & None \\
Map-free (Match)~\cite{arnold2022map} $\dagger$ & 9.09 / 5.33 & 2.51 / 3.11 & 3.89 / 6.44 & 1.04 / 3.61 & 3.00 / 6.14 & 2.61 / 4.83 & 3.90 / 4.93 & None \\
Map-free (Regress)~\cite{arnold2022map} & 8.40 / 4.56 & 2.44 / 2.54 & 3.73 / 5.23 & 0.97 / 3.17 & 2.91 / 5.10 & 2.51 / 4.01 & 3.69 / 4.12 & None \\
ExReNet (SN)~\cite{ExReNet} & 10.97 / 6.52 & 2.34 / 2.99 & 3.40 / 4.92 & 0.90 / 3.23 & 2.60 / 4.69 & 2.36 / 4.23 & 4.08 / 4.32 & None \\
ExReNet (SUNCG)~\cite{ExReNet} & 9.79 / 4.46 & 2.33 / 2.48 & 3.54 / 3.49 & 0.72 / 2.41 & 2.30 / 3.74 & 2.22 / 3.03 & 3.74 / 3.31 & None \\
ImageNet+NCM~\cite{zhou2020learn} $\dagger$ & -- & -- & -- & -- & -- & 1.83 / 0.56 & -- & None \\
Reloc3r-512~\cite{reloc3r} & 1.22 / 0.73 & 0.42 / 0.36 & 0.62 / 0.55 & 0.13 / 0.58 & 0.34 / 0.58 & 0.38 / 0.52 & 0.55 / 0.56 & None \\
\textbf{Reloc-VGGT(mask)} & 0.98 / 0.63 & 0.40 / 0.48 & 0.59 / 0.52 & 0.14 / 0.56 & 0.31 / 0.55 & 0.36 / 0.55 & 0.48 / 0.55 & None \\
\textbf{Reloc-VGGT} & \textbf{0.59} / \textbf{0.43} & \textbf{0.36} / \textbf{0.36} & \textbf{0.57} / \textbf{0.44} & \textbf{0.10} / \textbf{0.33} & \textbf{0.24} / \textbf{0.43} & \textbf{0.32} / \textbf{0.37} & \textbf{0.37} / \textbf{0.38} & None \\
\bottomrule
\end{tabular}}}
\caption{Pose estimation results (median translation / rotation error) on the Cambridge Landmarks~\cite{cambridge} dataset. APR: absolute pose regression; RPR: relative pose regression.}
\label{tab:cambridge}
\vspace{-0.3cm}
\end{table*}

\begin{table*}[t]
\centering

\setlength{\tabcolsep}{6pt} 
\begin{tabular}{lccc}
\toprule
\textbf{Model Variant} & \textbf{7-Scenes (Avg.)} & \textbf{Cambridge Landmarks (Avg.)} & \textbf{Inference Time (s)} \\
                       & \textit{Trans. (m) / Rot. (°)} & \textit{Trans. (m) / Rot. (°)} & \\
\midrule
a. w/o RPTP \& SMA (VGGT) & 0.034/0.908 & 0.39/0.40 & 4.74        \\
b. w/o RPTP & 0.051/1.241 & 0.64/1.16 & 3.10        \\
c. w/o SMA & 0.031/0.896 & 0.37/0.38 & 4.82       \\ 
d. w RPTP \& SMA & 0.039/1.033 & 0.48/0.55 & 3.14       \\
\bottomrule
\end{tabular}
\vspace{-0.3cm}
\caption{Ablation Study on Core Components of Reloc-VGGT. RPTP: Relative Pose Tokenizer and Projection. SMA: Sparse Mask Attention.}
\label{tab:ablation_core}
\vspace{-0.3cm}
\end{table*}

\begin{table}[htbp]
\centering
\scalebox{0.8}{
\setlength{\tabcolsep}{0.9mm}{
\begin{tabular}{lcccc}
\hline
\textbf{Mask type} & \textbf{chess} & \textbf{fire} & \textbf{heads} & \textbf{Inference time} \\
\hline
Global Attention & 0.025/0.767 & 0.024/0.833 & 0.013/0.770 & 4.82 \\
Causal Attention & 0.039/0.898 & 0.034/1.013 & 0.019/0.894 & 4.71 \\
Masked Attention & 0.027/0.861 & 0.028/0.871 & 0.014/0.818 & 3.14 \\
Dilated Attention & 0.026/0.837 & 0.027/0.864 & 0.013/0.809 & 4.03 \\
\hline
\end{tabular}}}
\vspace{-0.2cm}
\caption{Pose estimation accuracy (translation error (m) / rotation error (°)) and inference time per sequence(s) under different masking strategies on the 7-Scenes dataset.}
\label{tab:mask_performance_7scenes}
\vspace{-0.3cm}
\end{table}

\subsection{Ablation Studies}
To validate the effectiveness of each core component in Reloc-VGGT, we conducted systematic ablation experiments on the 7-Scenes and Cambridge Landmarks datasets. All experiments maintained identical configurations to ensure fair comparisons.  

First, we verified the necessity of the two proposed core modules: the relative Pose Tokenizer and Projection (RPTP) and Sparse Mask Attention (SMA). The experimental configurations were designed as follows:  
a. Retain the backbone network but remove the aforementioned two modules, and SMA with full global attention, the same with VGGT.  
b. Remove only the relative Pose Tokenizer and Projection (RPTP). 
c. Remove only the Sparse Mask Attention (SMA). 

The experimental results are presented in Tab.~\ref{tab:ablation_core}. It can be observed that after incorporating RPTP, the average translation/rotation errors are significantly reduced. This demonstrates that relative pose regression enables effective interaction between spatial information and visual features, thereby enhancing the accuracy of pose estimation. When SMA is used alone, the inference time is reduced by 35\% compared to the baseline model, with only minor performance degradation. This validates that sparse attention with linear complexity can efficiently reduce redundant computations without losing critical geometric context. The full model achieves an optimal balance between accuracy and efficiency.  
In Tab.~\ref{tab:mask_performance_7scenes}, we evaluate the impact of different mask types on final performance. 
We observe that, although the sparse mask attention incurs a slight accuracy drop compared with the dilated mask, it offers significantly better real-time performance. 
In contrast, the causal mask performs poorly in both accuracy and speed. 
These results demonstrate that our sparse mask design is well aligned with the characteristics of the visual localization task, as it prioritizes attention around the query pose token where information is most critical.

\begin{figure}[h]
    \centering
    \includegraphics[width=\linewidth]{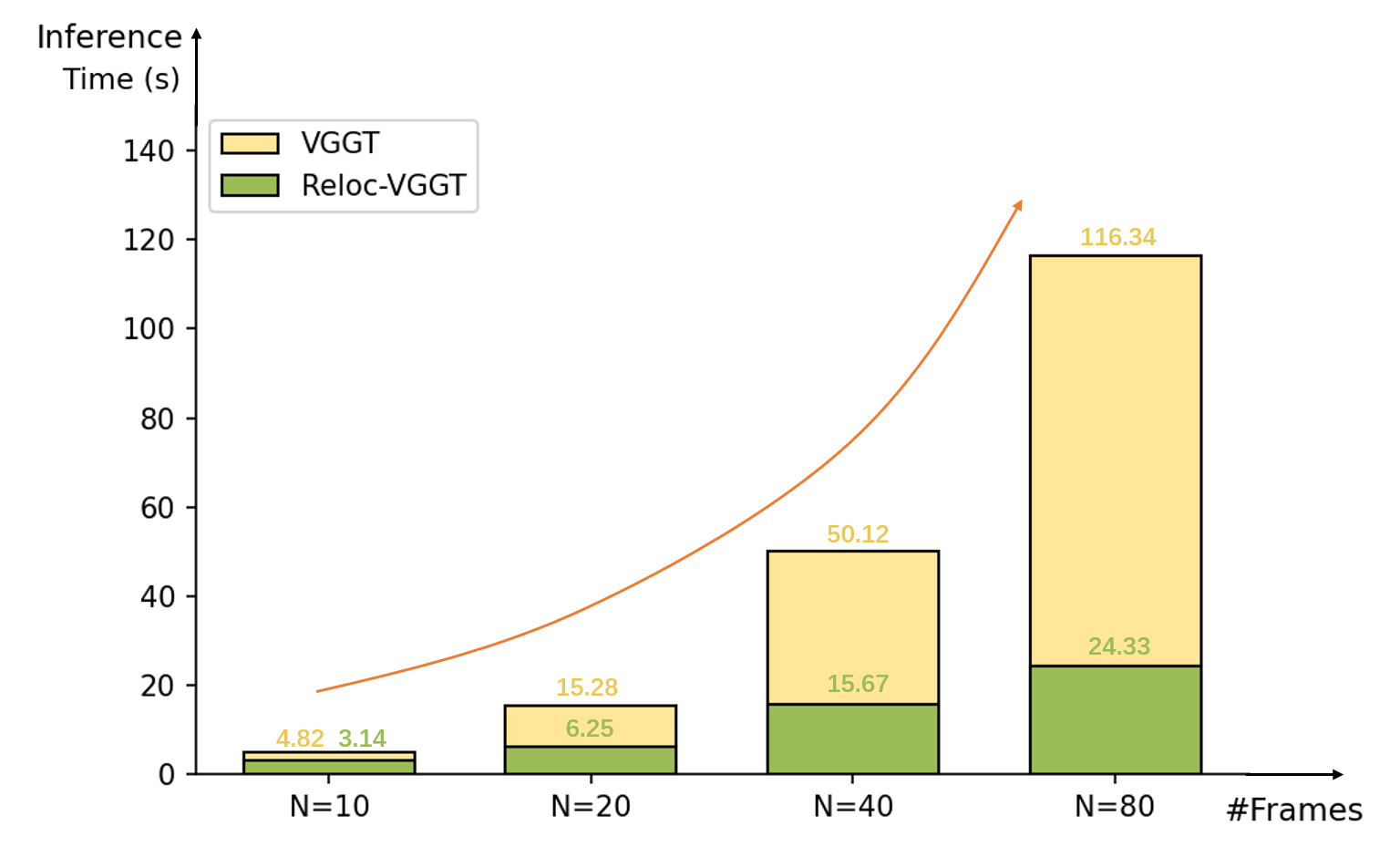}
    \vspace{-0.4cm}
    \caption{We visualize the runtime trend as the number of selected top-$k$ images increases, comparing the global attention mechanism with our proposed sparse mask attention. Our sparse mask attention strategy significantly reduces inference time, making relocalization over longer sequences computationally feasible.}
    \label{fig:topk}
    \vspace{-0.4cm}
\end{figure}

We further analyzed the impact of $K$ (the number of retrieved database frames) on performance, where $K$ ranges from 10 to 80, covering typical settings in visual relocalization tasks. The experimental results are shown in  Fig.~\ref{fig:topk}. It is evident that when $K$ increases, the visual localization accuracy improves. This is attributed to the richer spatial context provided by more source frames. However, when $K$ exceeds 20, accuracy gains saturate while inference time increases substantially. The original global attention mechanism in VGGT exhibits a quadratic growth with respect to the number of frames, while our method achieves linear-time scaling. This substantially improves the practicality of our framework and greatly enhances its deployment potential in real-world scenarios.

\subsection{Conclusion}
In this paper, we propose Reloc-VGGT, a novel visual re-localization framework with 3D foundation model. We introduce a pose tokenizer and projection module that effectively inject the spatial information from source frames into the attention blocks. Furthermore, we design a sparse mask attention mechanism that reduces the computational complexity from quadratic to linear, significantly enhancing real-time performance and scalability. Extensive experiments demonstrate that our method achieves superior performance in terms of generalization ability and pose estimation accuracy, highlighting the effectiveness of multi-view spatial integration for visual relocalization.

{
    \small
    \bibliographystyle{ieeenat_fullname}
    \bibliography{main}

@String(CVPR= {IEEE Conf. Comput. Vis. Pattern Recog.})

@String(ECCV= {Eur. Conf. Comput. Vis.})

@String(CVPR  = {CVPR})

@String(ECCV  = {ECCV})

@inproceedings{NeRF,
author = {Mildenhall, Ben and Srinivasan, Pratul P. and Tancik, Matthew and Barron, Jonathan T. and Ramamoorthi, Ravi and Ng, Ren},
title = {NeRF: Representing Scenes as Neural Radiance Fields for View Synthesis},
booktitle = {ECCV},
year = {2020},
}

@ARTICLE{orbslam,
  author={Mur-Artal, Raúl and Montiel, J. M. M. and Tardós, Juan D.},
  journal={IEEE Transactions on Robotics}, 
  title={ORB-SLAM: A Versatile and Accurate Monocular SLAM System}, 
  year={2015},
  volume={31},
  number={5},
  pages={1147-1163},
  doi={10.1109/TRO.2015.2463671}}

@ARTICLE{vins,
  author={Qin, Tong and Li, Peiliang and Shen, Shaojie},
  journal={IEEE Transactions on Robotics}, 
  title={VINS-Mono: A Robust and Versatile Monocular Visual-Inertial State Estimator}, 
  year={2018},
  volume={34},
  number={4},
  pages={1004-1020},
  doi={10.1109/TRO.2018.2853729}}

@INPROCEEDINGS{inerf,
  author={Yen-Chen, Lin and Florence, Pete and Barron, Jonathan T. and Rodriguez, Alberto and Isola, Phillip and Lin, Tsung-Yi},
  booktitle={2021 IEEE/RSJ International Conference on Intelligent Robots and Systems (IROS)}, 
  title={iNeRF: Inverting Neural Radiance Fields for Pose Estimation}, 
  year={2021},
  volume={},
  number={},
  pages={1323-1330},
  doi={10.1109/IROS51168.2021.9636708}}

@article{nerf--,
  title={NeRF--: Neural radiance fields without known camera parameters},
  author={Wang, Zirui and Wu, Shangzhe and Xie, Weidi and Chen, Min and Prisacariu, Victor Adrian},
  journal={arXiv preprint arXiv:2102.07064},
  year={2021}
}

@InProceedings{scannet,
author = {Dai, Angela and Chang, Angel X. and Savva, Manolis and Halber, Maciej and Funkhouser, Thomas and Niessner, Matthias},
title = {ScanNet: Richly-Annotated 3D Reconstructions of Indoor Scenes},
booktitle = {Proceedings of the IEEE Conference on Computer Vision and Pattern Recognition (CVPR)},
month = {July},
year = {2017}
}

@InProceedings{superpoint,
author = {DeTone, Daniel and Malisiewicz, Tomasz and Rabinovich, Andrew},
title = {SuperPoint: Self-Supervised Interest Point Detection and Description},
booktitle = {Proceedings of the IEEE Conference on Computer Vision and Pattern Recognition (CVPR) Workshops},
month = {June},
year = {2018}
}

@article{prosgnerf,
  title={ProSGNeRF: Progressive Dynamic Neural Scene Graph with Frequency Modulated Auto-Encoder in Urban Scenes},
  author={Deng, Tianchen and Liu, Siyang and Wang, Xuan and Liu, Yejia and Wang, Danwei and Chen, Weidong},
  journal={arXiv preprint arXiv:2312.09076},
  year={2023}
}

@inproceedings{plgslam,
  title={Plgslam: Progressive neural scene represenation with local to global bundle adjustment},
  author={Deng, Tianchen and Shen, Guole and Qin, Tong and Wang, Jianyu and Zhao, Wentao and Wang, Jingchuan and Wang, Danwei and Chen, Weidong},
  booktitle={Proceedings of the IEEE/CVF Conference on Computer Vision and Pattern Recognition},
  pages={19657--19666},
  year={2024}
}

@article{3dgs,
  title={3D Gaussian Splatting for Real-Time Radiance Field Rendering.},
  author={Kerbl, Bernhard and Kopanas, Georgios and Leimk{\"u}hler, Thomas and Drettakis, George},
  journal={ACM Trans. Graph.},
  volume={42},
  number={4},
  pages={139--1},
  year={2023}
}

@article{oxfordrobo,
  title={1 year, 1000 km: The oxford robotcar dataset},
  author={Maddern, Will and Pascoe, Geoffrey and Linegar, Chris and Newman, Paul},
  journal={The International Journal of Robotics Research},
  volume={36},
  number={1},
  pages={3--15},
  year={2017},
  publisher={SAGE Publications Sage UK: London, England}
}

@ARTICLE{neslam,
  author={Deng, Tianchen and Wang, Yanbo and Xie, Hongle and Wang, Hesheng and Guo, Rui and Wang, Jingchuan and Wang, Danwei and Chen, Weidong},
  journal={IEEE Transactions on Automation Science and Engineering}, 
  title={NeSLAM: Neural Implicit Mapping and Self-Supervised Feature Tracking With Depth Completion and Denoising}, 
  year={2025},
  volume={},
  number={},
  pages={1-1},
  doi={10.1109/TASE.2025.3541064}}

@ARTICLE{incremental,
  author={Deng, Tianchen and Wang, Nailin and Wang, Chongdi and Yuan, Shenghai and Wang, Jingchuan and Wang, Hesheng and Wang, Danwei and Chen, Weidong},
  journal={IEEE Transactions on Automation Science and Engineering}, 
  title={Incremental Joint Learning of Depth, Pose and Implicit Scene Representation on Monocular Camera in Large-scale Scenes}, 
  year={2025},
  volume={},
  number={},
  pages={1-1},
  doi={10.1109/TASE.2025.3617654}}

@article{deng2025vpgs,
  title={VPGS-SLAM: Voxel-based Progressive 3D Gaussian SLAM in Large-Scale Scenes},
  author={Deng, Tianchen and Wu, Wenhua and He, Junjie and Pan, Yue and Jiang, Xirui and Yuan, Shenghai and Wang, Danwei and Wang, Hesheng and Chen, Weidong},
  journal={arXiv preprint arXiv:2505.18992},
  year={2025}
}

@article{deng2025mcnslam,
      title={MCN-SLAM: Multi-Agent Collaborative Neural SLAM with Hybrid Implicit Neural Scene Representation}, 
      author={Tianchen Deng and Guole Shen and Xun Chen and Shenghai Yuan and Hongming Shen and Guohao Peng and Zhenyu Wu and Jingchuan Wang and Lihua Xie and Danwei Wang and Hesheng Wang and Weidong Chen},
      journal={arXiv preprint arXiv:2506.18678},
      year={2025},
}

@article{grsslam,
  title={GRS-SLAM3R: Real-Time Dense SLAM with Gated Recurrent State},
  author={Shen, Guole and Deng, Tianchen and Wang, Yanbo and Chen, Yongtao and Shen, Yilin and Liu, Jiuming and Wang, Jingchuan},
  journal={arXiv preprint arXiv:2509.23737},
  year={2025}
}

@article{lenav,
  title={Learning to Tune Like an Expert: Interpretable and Scene-Aware Navigation via MLLM Reasoning and CVAE-Based Adaptation},
  author={Wang, Yanbo and Fang, Zipeng and Zhao, Lei and Chen, Weidong},
  journal={arXiv preprint arXiv:2507.11001},
  year={2025}
}

@article{compactslam,
  title={Compact 3D Gaussian Splatting For Dense Visual SLAM},
  author={Deng, Tianchen and Chen, Yaohui and Zhang, Leyan and Yang, Jianfei and Yuan, Shenghai and Wang, Danwei and Chen, Weidong},
  journal={arXiv preprint arXiv:2403.11247},
  year={2024}
}

@article{mgslam,
  title={MG-SLAM: Structure Gaussian Splatting SLAM with Manhattan World Hypothesis},
  author={Liu, Shuhong and Deng, Tianchen and Zhou, Heng and Li, Liuzhuozheng and Wang, Hongyu and Wang, Danwei and Li, Mingrui},
  journal={IEEE Transactions on Automation Science and Engineering},
  year={2025},
  publisher={IEEE}
}

@inproceedings{mneslam,
  title={Mne-slam: Multi-agent neural slam for mobile robots},
  author={Deng, Tianchen and Shen, Guole and Xun, Chen and Yuan, Shenghai and Jin, Tongxin and Shen, Hongming and Wang, Yanbo and Wang, Jingchuan and Wang, Hesheng and Wang, Danwei and others},
  booktitle={Proceedings of the Computer Vision and Pattern Recognition Conference},
  pages={1485--1494},
  year={2025}
}

@inproceedings{vggt,
  title={Vggt: Visual geometry grounded transformer},
  author={Wang, Jianyuan and Chen, Minghao and Karaev, Nikita and Vedaldi, Andrea and Rupprecht, Christian and Novotny, David},
  booktitle={Proceedings of the Computer Vision and Pattern Recognition Conference},
  pages={5294--5306},
  year={2025}
}

@inproceedings{netvlad,
  title={NetVLAD: CNN architecture for weakly supervised place recognition},
  author={Arandjelovic, Relja and Gronat, Petr and Torii, Akihiko and Pajdla, Tomas and Sivic, Josef},
  booktitle={Proceedings of the IEEE conference on computer vision and pattern recognition},
  pages={5297--5307},
  year={2016}
}

@article{dinov2,
  title={Dinov2: Learning robust visual features without supervision},
  author={Oquab, Maxime and Darcet, Timoth{\'e}e and Moutakanni, Th{\'e}o and Vo, Huy and Szafraniec, Marc and Khalidov, Vasil and Fernandez, Pierre and Haziza, Daniel and Massa, Francisco and El-Nouby, Alaaeldin and others},
  journal={arXiv preprint arXiv:2304.07193},
  year={2023}
}

@inproceedings{superglue,
  title={Superglue: Learning feature matching with graph neural networks},
  author={Sarlin, Paul-Edouard and DeTone, Daniel and Malisiewicz, Tomasz and Rabinovich, Andrew},
  booktitle={Proceedings of the IEEE/CVF conference on computer vision and pattern recognition},
  pages={4938--4947},
  year={2020}
}

@inproceedings{brachmann2018learning,
  title={Learning less is more-6d camera localization via 3d surface regression},
  author={Brachmann, Eric and Rother, Carsten},
  booktitle={Proceedings of the IEEE conference on computer vision and pattern recognition},
  pages={4654--4662},
  year={2018}
}

@inproceedings{reloc3r,
  title={Reloc3r: Large-scale training of relative camera pose regression for generalizable, fast, and accurate visual localization},
  author={Dong, Siyan and Wang, Shuzhe and Liu, Shaohui and Cai, Lulu and Fan, Qingnan and Kannala, Juho and Yang, Yanchao},
  booktitle={Proceedings of the Computer Vision and Pattern Recognition Conference},
  pages={16739--16752},
  year={2025}
}

@article{brachmann2021visual,
  title={Visual camera re-localization from RGB and RGB-D images using DSAC},
  author={Brachmann, Eric and Rother, Carsten},
  journal={IEEE transactions on pattern analysis and machine intelligence},
  volume={44},
  number={9},
  pages={5847--5865},
  year={2021},
  publisher={IEEE}
}

@inproceedings{brachmann2023accelerated,
  title={Accelerated coordinate encoding: Learning to relocalize in minutes using rgb and poses},
  author={Brachmann, Eric and Cavallari, Tommaso and Prisacariu, Victor Adrian},
  booktitle={Proceedings of the IEEE/CVF Conference on Computer Vision and Pattern Recognition},
  pages={5044--5053},
  year={2023}
}

@inproceedings{li2020hierarchical,
  title={Hierarchical scene coordinate classification and regression for visual localization},
  author={Li, Xiaotian and Wang, Shuzhe and Zhao, Yi and Verbeek, Jakob and Kannala, Juho},
  booktitle={Proceedings of the IEEE/CVF Conference on Computer Vision and Pattern Recognition},
  pages={11983--11992},
  year={2020}
}

@inproceedings{brahmbhatt2018geometry,
  title={Geometry-aware learning of maps for camera localization},
  author={Brahmbhatt, Samarth and Gu, Jinwei and Kim, Kihwan and Hays, James and Kautz, Jan},
  booktitle={Proceedings of the IEEE conference on computer vision and pattern recognition},
  pages={2616--2625},
  year={2018}
}

@inproceedings{kendall2017geometric,
  title={Geometric loss functions for camera pose regression with deep learning},
  author={Kendall, Alex and Cipolla, Roberto},
  booktitle={Proceedings of the IEEE conference on computer vision and pattern recognition},
  pages={5974--5983},
  year={2017}
}

@inproceedings{shavit2021learning,
  title={Learning multi-scene absolute pose regression with transformers},
  author={Shavit, Yoli and Ferens, Ron and Keller, Yosi},
  booktitle={Proceedings of the IEEE/CVF International Conference on Computer Vision},
  pages={2733--2742},
  year={2021}
}

@article{shavit2023coarse,
  title={Coarse-to-fine multi-scene pose regression with transformers},
  author={Shavit, Yoli and Ferens, Ron and Keller, Yosi},
  journal={IEEE transactions on pattern analysis and machine intelligence},
  volume={45},
  number={12},
  pages={14222--14233},
  year={2023},
  publisher={IEEE}
}

@inproceedings{walch2017image,
  title={Image-based localization using lstms for structured feature correlation},
  author={Walch, Florian and Hazirbas, Caner and Leal-Taixe, Laura and Sattler, Torsten and Hilsenbeck, Sebastian and Cremers, Daniel},
  booktitle={Proceedings of the IEEE international conference on computer vision},
  pages={627--637},
  year={2017}
}

@inproceedings{arnold2022map,
  title={Map-free visual relocalization: Metric pose relative to a single image},
  author={Arnold, Eduardo and Wynn, Jamie and Vicente, Sara and Garcia-Hernando, Guillermo and Monszpart, Aron and Prisacariu, Victor and Turmukhambetov, Daniyar and Brachmann, Eric},
  booktitle={European Conference on Computer Vision},
  pages={690--708},
  year={2022},
  organization={Springer}
}

@inproceedings{balntas2018relocnet,
  title={Relocnet: Continuous metric learning relocalisation using neural nets},
  author={Balntas, Vassileios and Li, Shuda and Prisacariu, Victor},
  booktitle={Proceedings of the European conference on computer vision (ECCV)},
  pages={751--767},
  year={2018}
}

@inproceedings{zhou2020learn,
  title={To learn or not to learn: Visual localization from essential matrices},
  author={Zhou, Qunjie and Sattler, Torsten and Pollefeys, Marc and Leal-Taixe, Laura},
  booktitle={2020 IEEE International Conference on Robotics and Automation (ICRA)},
  pages={3319--3326},
  year={2020},
  organization={IEEE}
}

@inproceedings{ding2019camnet,
  title={CamNet: Coarse-to-fine retrieval for camera re-localization},
  author={Ding, Mingyu and Wang, Zhe and Sun, Jiankai and Shi, Jianping and Luo, Ping},
  booktitle={Proceedings of the IEEE/CVF International Conference on Computer Vision},
  pages={2871--2880},
  year={2019}
}

@inproceedings{dust3r,
  title={Dust3r: Geometric 3d vision made easy},
  author={Wang, Shuzhe and Leroy, Vincent and Cabon, Yohann and Chidlovskii, Boris and Revaud, Jerome},
  booktitle={Proceedings of the IEEE/CVF Conference on Computer Vision and Pattern Recognition},
  pages={20697--20709},
  year={2024}
}

@inproceedings{chen2024leveraging,
  title={Leveraging neural radiance fields for uncertainty-aware visual localization},
  author={Chen, Le and Chen, Weirong and Wang, Rui and Pollefeys, Marc},
  booktitle={2024 IEEE International Conference on Robotics and Automation (ICRA)},
  pages={6298--6305},
  year={2024},
  organization={IEEE}
}

@article{zhai2025splatloc,
  title={Splatloc: 3d gaussian splatting-based visual localization for augmented reality},
  author={Zhai, Hongjia and Zhang, Xiyu and Zhao, Boming and Li, Hai and He, Yijia and Cui, Zhaopeng and Bao, Hujun and Zhang, Guofeng},
  journal={IEEE Transactions on Visualization and Computer Graphics},
  year={2025},
  publisher={IEEE}
}

@inproceedings{kendall2015posenet,
  title={Posenet: A convolutional network for real-time 6-dof camera relocalization},
  author={Kendall, Alex and Grimes, Matthew and Cipolla, Roberto},
  booktitle={Proceedings of the IEEE international conference on computer vision},
  pages={2938--2946},
  year={2015}
}

@inproceedings{clark2017vidloc,
  title={Vidloc: A deep spatio-temporal model for 6-dof video-clip relocalization},
  author={Clark, Ronald and Wang, Sen and Markham, Andrew and Trigoni, Niki and Wen, Hongkai},
  booktitle={Proceedings of the IEEE conference on computer vision and pattern recognition},
  pages={6856--6864},
  year={2017}
}

@inproceedings{xue2019local,
  title={Local supports global: Deep camera relocalization with sequence enhancement},
  author={Xue, Fei and Wang, Xin and Yan, Zike and Wang, Qiuyuan and Wang, Junqiu and Zha, Hongbin},
  booktitle={Proceedings of the IEEE/CVF International Conference on Computer Vision},
  pages={2841--2850},
  year={2019}
}

@inproceedings{schonberger2016structure,
  title={Structure-from-motion revisited},
  author={Schonberger, Johannes L and Frahm, Jan-Michael},
  booktitle={Proceedings of the IEEE conference on computer vision and pattern recognition},
  pages={4104--4113},
  year={2016}
}

@inproceedings{turkoglu2021visual,
  title={Visual camera re-localization using graph neural networks and relative pose supervision},
  author={Turkoglu, Mehmet Ozgur and Brachmann, Eric and Schindler, Konrad and Brostow, Gabriel J and Monszpart, Aron},
  booktitle={2021 International Conference on 3D Vision (3DV)},
  pages={145--155},
  year={2021},
  organization={IEEE}
}

@article{vaswani2017attention,
  title={Attention is all you need},
  author={Vaswani, Ashish and Shazeer, Noam and Parmar, Niki and Uszkoreit, Jakob and Jones, Llion and Gomez, Aidan N and Kaiser, {\L}ukasz and Polosukhin, Illia},
  journal={Advances in neural information processing systems},
  volume={30},
  year={2017}
}

@inproceedings{bai2024sequential,
  title={Sequential modeling enables scalable learning for large vision models},
  author={Bai, Yutong and Geng, Xinyang and Mangalam, Karttikeya and Bar, Amir and Yuille, Alan L and Darrell, Trevor and Malik, Jitendra and Efros, Alexei A},
  booktitle={Proceedings of the IEEE/CVF Conference on Computer Vision and Pattern Recognition},
  pages={22861--22872},
  year={2024}
}

@inproceedings{kirillov2023segment,
  title={Segment anything},
  author={Kirillov, Alexander and Mintun, Eric and Ravi, Nikhila and Mao, Hanzi and Rolland, Chloe and Gustafson, Laura and Xiao, Tete and Whitehead, Spencer and Berg, Alexander C and Lo, Wan-Yen and others},
  booktitle={Proceedings of the IEEE/CVF international conference on computer vision},
  pages={4015--4026},
  year={2023}
}

@inproceedings{peebles2023scalable,
  title={Scalable diffusion models with transformers},
  author={Peebles, William and Xie, Saining},
  booktitle={Proceedings of the IEEE/CVF international conference on computer vision},
  pages={4195--4205},
  year={2023}
}

@article{bai2023qwen,
  title={Qwen technical report},
  author={Bai, Jinze and Bai, Shuai and Chu, Yunfei and Cui, Zeyu and Dang, Kai and Deng, Xiaodong and Fan, Yang and Ge, Wenbin and Han, Yu and Huang, Fei and others},
  journal={arXiv preprint arXiv:2309.16609},
  year={2023}
}

@article{chowdhery2023palm,
  title={Palm: Scaling language modeling with pathways},
  author={Chowdhery, Aakanksha and Narang, Sharan and Devlin, Jacob and Bosma, Maarten and Mishra, Gaurav and Roberts, Adam and Barham, Paul and Chung, Hyung Won and Sutton, Charles and Gehrmann, Sebastian and others},
  journal={Journal of Machine Learning Research},
  volume={24},
  number={240},
  pages={1--113},
  year={2023}
}

@article{minaee2024large,
  title={Large language models: A survey},
  author={Minaee, Shervin and Mikolov, Tomas and Nikzad, Narjes and Chenaghlu, Meysam and Socher, Richard and Amatriain, Xavier and Gao, Jianfeng},
  journal={arXiv preprint arXiv:2402.06196},
  year={2024}
}

@inproceedings{cabon2025must3r,
  title={Must3r: Multi-view network for stereo 3d reconstruction},
  author={Cabon, Yohann and Stoffl, Lucas and Antsfeld, Leonid and Csurka, Gabriela and Chidlovskii, Boris and Revaud, Jerome and Leroy, Vincent},
  booktitle={Proceedings of the Computer Vision and Pattern Recognition Conference},
  pages={1050--1060},
  year={2025}
}

@article{fastvggt,
  title={Fastvggt: Training-free acceleration of visual geometry transformer},
  author={Shen, You and Zhang, Zhipeng and Qu, Yansong and Cao, Liujuan},
  journal={arXiv preprint arXiv:2509.02560},
  year={2025}
}

@article{streamvggt,
  title={Streaming 4d visual geometry transformer},
  author={Zhuo, Dong and Zheng, Wenzhao and Guo, Jiahe and Wu, Yuqi and Zhou, Jie and Lu, Jiwen},
  journal={arXiv preprint arXiv:2507.11539},
  year={2025}
}

@inproceedings{mast3r,
  title={Grounding image matching in 3d with mast3r},
  author={Leroy, Vincent and Cabon, Yohann and Revaud, J{\'e}r{\^o}me},
  booktitle={European Conference on Computer Vision},
  pages={71--91},
  year={2024},
  organization={Springer}
}

@inproceedings{co3dv2,
  title={Common objects in 3d: Large-scale learning and evaluation of real-life 3d category reconstruction},
  author={Reizenstein, Jeremy and Shapovalov, Roman and Henzler, Philipp and Sbordone, Luca and Labatut, Patrick and Novotny, David},
  booktitle={Proceedings of the IEEE/CVF international conference on computer vision},
  pages={10901--10911},
  year={2021}
}

@inproceedings{LENS,
  title={Lens: Localization enhanced by nerf synthesis},
  author={Moreau, Arthur and Piasco, Nathan and Tsishkou, Dzmitry and Stanciulescu, Bogdan and de La Fortelle, Arnaud},
  booktitle={Conference on Robot Learning},
  pages={1347--1356},
  year={2022},
  organization={PMLR}
}

@inproceedings{PMNet,
  title={Learning neural volumetric pose features for camera localization},
  author={Lin, Jingyu and Gu, Jiaqi and Wu, Bojian and Fan, Lubin and Chen, Renjie and Liu, Ligang and Ye, Jieping},
  booktitle={European Conference on Computer Vision},
  pages={198--214},
  year={2024},
  organization={Springer}
}

@inproceedings{DFNet,
  title={Dfnet: Enhance absolute pose regression with direct feature matching},
  author={Chen, Shuai and Li, Xinghui and Wang, Zirui and Prisacariu, Victor A},
  booktitle={European Conference on Computer Vision},
  pages={1--17},
  year={2022},
  organization={Springer}
}

@inproceedings{NeFeS,
  title={Neural refinement for absolute pose regression with feature synthesis},
  author={Chen, Shuai and Bhalgat, Yash and Li, Xinghui and Bian, Jia-Wang and Li, Kejie and Wang, Zirui and Prisacariu, Victor Adrian},
  booktitle={Proceedings of the IEEE/CVF Conference on Computer Vision and Pattern Recognition},
  pages={20987--20996},
  year={2024}
}

@inproceedings{Marepo,
  title={Map-relative pose regression for visual re-localization},
  author={Chen, Shuai and Cavallari, Tommaso and Prisacariu, Victor Adrian and Brachmann, Eric},
  booktitle={Proceedings of the IEEE/CVF Conference on Computer Vision and Pattern Recognition},
  pages={20665--20674},
  year={2024}
}

@inproceedings{RelativePN,
  title={Camera relocalization by computing pairwise relative poses using convolutional neural network},
  author={Laskar, Zakaria and Melekhov, Iaroslav and Kalia, Surya and Kannala, Juho},
  booktitle={Proceedings of the IEEE international conference on computer vision workshops},
  pages={929--938},
  year={2017}
}

@article{anchornet,
  title={Improved visual relocalization by discovering anchor points},
  author={Saha, Soham and Varma, Girish and Jawahar, CV},
  journal={arXiv preprint arXiv:1811.04370},
  year={2018}
}

@inproceedings{ExReNet,
  title={Learning to localize in new environments from synthetic training data},
  author={Winkelbauer, Dominik and Denninger, Maximilian and Triebel, Rudolph},
  booktitle={2021 IEEE International Conference on Robotics and Automation (ICRA)},
  pages={5840--5846},
  year={2021},
  organization={IEEE}
}

@article{flashattention,
  title={Flashattention-2: Faster attention with better parallelism and work partitioning},
  author={Dao, Tri},
  journal={arXiv preprint arXiv:2307.08691},
  year={2023}
}

@inproceedings{7scenes,
  title={Scene coordinate regression forests for camera relocalization in RGB-D images},
  author={Shotton, Jamie and Glocker, Ben and Zach, Christopher and Izadi, Shahram and Criminisi, Antonio and Fitzgibbon, Andrew},
  booktitle={Proceedings of the IEEE conference on computer vision and pattern recognition},
  pages={2930--2937},
  year={2013}
}

@inproceedings{cambridge,
  title={Posenet: A convolutional network for real-time 6-dof camera relocalization},
  author={Kendall, Alex and Grimes, Matthew and Cipolla, Roberto},
  booktitle={Proceedings of the IEEE international conference on computer vision},
  pages={2938--2946},
  year={2015}
}

@inproceedings{LoFTR,
  title={Efficient LoFTR: Semi-dense local feature matching with sparse-like speed},
  author={Wang, Yifan and He, Xingyi and Peng, Sida and Tan, Dongli and Zhou, Xiaowei},
  booktitle={Proceedings of the IEEE/CVF conference on computer vision and pattern recognition},
  pages={21666--21675},
  year={2024}
}

@inproceedings{roma,
  title={Roma: Robust dense feature matching},
  author={Edstedt, Johan and Sun, Qiyu and B{\"o}kman, Georg and Wadenb{\"a}ck, M{\aa}rten and Felsberg, Michael},
  booktitle={Proceedings of the IEEE/CVF Conference on Computer Vision and Pattern Recognition},
  pages={19790--19800},
  year={2024}
}

@article{noposplat,
  title={No pose, no problem: Surprisingly simple 3d gaussian splats from sparse unposed images},
  author={Ye, Botao and Liu, Sifei and Xu, Haofei and Li, Xueting and Pollefeys, Marc and Yang, Ming-Hsuan and Peng, Songyou},
  journal={arXiv preprint arXiv:2410.24207},
  year={2024}
}

@inproceedings{posediffusion,
  title={Posediffusion: Solving pose estimation via diffusion-aided bundle adjustment},
  author={Wang, Jianyuan and Rupprecht, Christian and Novotny, David},
  booktitle={Proceedings of the IEEE/CVF International Conference on Computer Vision},
  pages={9773--9783},
  year={2023}
}

@inproceedings{pixsfm,
  title={Pixel-perfect structure-from-motion with featuremetric refinement},
  author={Lindenberger, Philipp and Sarlin, Paul-Edouard and Larsson, Viktor and Pollefeys, Marc},
  booktitle={Proceedings of the IEEE/CVF international conference on computer vision},
  pages={5987--5997},
  year={2021}
}

@inproceedings{relpose,
  title={Relpose: Predicting probabilistic relative rotation for single objects in the wild},
  author={Zhang, Jason Y and Ramanan, Deva and Tulsiani, Shubham},
  booktitle={European Conference on Computer Vision},
  pages={592--611},
  year={2022},
  organization={Springer}
}

@inproceedings{relposepp,
  title={Relpose++: Recovering 6d poses from sparse-view observations},
  author={Lin, Amy and Zhang, Jason Y and Ramanan, Deva and Tulsiani, Shubham},
  booktitle={2024 International Conference on 3D Vision (3DV)},
  pages={106--115},
  year={2024},
  organization={IEEE}
}

@article{raydiffusion,
  title={Cameras as rays: Pose estimation via ray diffusion},
  author={Zhang, Jason Y and Lin, Amy and Kumar, Moneish and Yang, Tzu-Hsuan and Ramanan, Deva and Tulsiani, Shubham},
  journal={arXiv preprint arXiv:2402.14817},
  year={2024}
}

@inproceedings{vggsfm,
  title={Vggsfm: Visual geometry grounded deep structure from motion},
  author={Wang, Jianyuan and Karaev, Nikita and Rupprecht, Christian and Novotny, David},
  booktitle={Proceedings of the IEEE/CVF conference on computer vision and pattern recognition},
  pages={21686--21697},
  year={2024}
}

@article{unipr-3d,
      title={UniPR-3D: Towards Universal Visual Place Recognition with Visual Geometry Grounded Transformer}, 
      author={Tianchen Deng and Xun Chen and Ziming Li and Hongming Shen and Danwei Wang and Javier Civera and Hesheng Wang},
      year={2025},
     journal={arXiv preprint 2512.21078}, 
}

@article{deng2025best3dscenerepresentation,
      title={What Is The Best 3D Scene Representation for Robotics? From Geometric to Foundation Models}, 
      author={Tianchen Deng and Yue Pan and Shenghai Yuan and Dong Li and Chen Wang and Mingrui Li and Long Chen and Lihua Xie and Danwei Wang and Jingchuan Wang and Javier Civera and Hesheng Wang and Weidong Chen},
      year={2025},
      journal={arXiv preprint arXiv:2512.03422}, 
}

@article{unilgl,
  title={UniLGL: Learning Uniform Place Recognition for FOV-limited/Panoramic LiDAR Global Localization},
  author={Shen, Hongming and Chen, Xun and Hui, Yulin and Wu, Zhenyu and Wang, Wei and Lyu, Qiyang and Deng, Tianchen and Wang, Danwei},
  journal={arXiv preprint arXiv:2507.12194},
  year={2025}
}
}


\end{document}